\documentclass{article}

    \usepackage[final]{neurips_2023}

\usepackage[utf8]{inputenc} 
\usepackage[T1]{fontenc}    
\usepackage{hyperref}       
\usepackage{url}            
\usepackage{booktabs}       
\usepackage{amsfonts}       
\usepackage{nicefrac}       
\usepackage{microtype}      
\usepackage{xcolor}         
\usepackage{amsmath}

\usepackage{graphicx}
\usepackage{color,soul}
\usepackage{placeins}
\usepackage{xspace}

\newcommand{\eg}{\emph{e.g.}\xspace}
\newcommand{\ie}{\emph{i.e.}\xspace}

\title{Understanding Representations Pretrained with Auxiliary Losses for Embodied Agent Planning}

\author{%
  Yuxuan Li \thanks{Work done while interning at AI2.} \\
  Stanford University, Allen Institute for AI\\
  \texttt{liyuxuan@stanford.edu} \\
  \And
  Luca Weihs \\
  Allen Institute for AI \\
  \texttt{lucaw@allenai.org}
}

\begin{document}

\maketitle

\begin{abstract}
  Pretrained representations from large-scale vision models have boosted the performance of downstream embodied policy learning. We look to understand whether additional self-supervised pretraining on exploration trajectories can build on these general-purpose visual representations to better support embodied planning in realistic environments. We evaluated four common auxiliary losses in embodied AI, two hindsight-based losses, and a standard imitation learning loss, by pretraining the agent's visual compression module and state belief representations with each objective and using CLIP as a representative visual backbone. The learned representations are then frozen for downstream multi-step evaluation on two goal-directed tasks. Surprisingly, we find that imitation learning on these exploration trajectories out-performs all other auxiliary losses even despite the exploration trajectories being dissimilar from the downstream tasks. This suggests that imitation of exploration may be ``all you need'' for building powerful planning representations. Additionally, we find that popular auxiliary losses can benefit from simple modifications to improve their support for downstream planning ability.
\end{abstract}

\section{Introduction}

Models trained on large-scale image and video datasets are now serving as vision foundation models for a variety of downstream tasks~\citep{nair2022r3m, oquab2023dinov2, radford2021learning, radosavovic2023real}. The representations produced by these pretrained, and frozen, models have proven highly effective in enabling high performance in embodied agents, \eg in navigation and manipulation~\citep{khandelwal2022simple, majumdar2023we, nair2022r3m, radosavovic2023real}. While these visual representations are \textit{general} (usable for many possible downstream tasks), they do not provide \textit{effective} support for agent planning in embodied environments out of the box (\ie, without extensive task-specific training).  Furthermore, \citet{majumdar2023we} showed that existing vision foundation models are yet to provide a universally useful signal for policy learning across all common embodied tasks; for example, generalization effectiveness can often depend on how in-domain the downstream observations are to the pretraining datasets. \citet{sharma2023lossless} suggests that the best downstream performance requires that the pretrained visual representations be further adapted during policy learning, \eg using reinforcement learning (RL) losses.

How can we get from generally-useful visual representations to effective representations for embodied planning, without direct training on the task objectives? Previously, researchers have shown that combining auxiliary tasks such as inverse dynamics prediction with reinforcement learning can increase performance, improve sample efficiency, and sometimes even lead to better learned representations~\citep{ye2021auxiliarypoint, ye2021auxiliaryobject, singh2022general}.  This suggests the possibility that, on top of a pretrained general-purpose visual backbone, pretraining agents on off-task trajectories using auxiliary objectives may help build effective embodied representations that generalize to downstream tasks.

In this work, we test the above hypothesis by pretraining agents on a fixed dataset of exploration trajectories and evaluate what unsupervised objectives best approximate direct on-task policy learning in leading to effective embodied representations.  We highlight three unique aspects of our approach. First, although evidence suggests that auxiliary losses can produce downstream improvements when co-trained with RL objectives, little is known about how they independently impact embodied representation learning.  We thus seek to disentangle their effects from RL, in addition to experimenting with novel unsupervised objectives.  Second, prior work has primarily focused on the effect of pretrained visual representations.  Here, we pretrain both the visual representations and the agent's belief representations in order to compare how different pretraining objectives can shape multi-frame, temporally dynamic representations. Third, to better understand how learned representations allow for longer-term embodied planning, we move from single-step to multi-step evaluation objectives and pretrain/evaluate the agent's learned representations in realistic-looking environments.

We report our initial set of experiments below, which explored common auxiliary losses used in embodied AI as well as two hindsight-based approaches that sought to encourage a form of unsupervised goal-directed pretraining. We pretrain both a convolutional visual compression module and a recurrent state encoder module on high-quality exploration trajectories, using an architecture similar to \cite{khandelwal2022simple}. These pretrained modules are then frozen in downstream evaluation, where the learned representations are evaluated for how well they transfer to multi-step imitation learning on two goal-directed tasks. For evaluation, we use both a standard object navigation task \citep[see][]{batra2020objectnav} and a novel navigation task in which the prior trajectory defines the agent's goal.  We emphasize that our goal is not to achieve the highest performance in these tasks, but rather to understand the relative performance of different pretraining objectives through a systematic comparison and a multi-step evaluation objective.  

Our preliminary results suggest that, although all pretraining objectives are more helpful than ``blind'' statistical baselines, those that require single-step action prediction generally lead to more generalizable representations in downstream multi-step evaluation. Surprisingly, simple behavior cloning on the pretraining expert exploration trajectories performed well in downstream evaluation, even though inverse dynamics and hindsight-based objectives were much more effective in achieving higher accuracy during pretraining. Moreover, we found that simple variants of common auxiliary losses are more helpful than their original formulations. These results suggest that the richness of behavior in the pretraining trajectories plays a significant role in shaping effective embodied representations that can transfer to different tasks, perhaps even more so than the exact pretraining objective used for agent representation learning.  This is in line with recent findings on the role of datasets in training powerful visual backbones~\citep{majumdar2023we}. We discuss the implications and challenges these results raise and future directions of our work.

\section{Methods}

\begin{figure}[h]
    \centering
    \includegraphics[width=14cm]{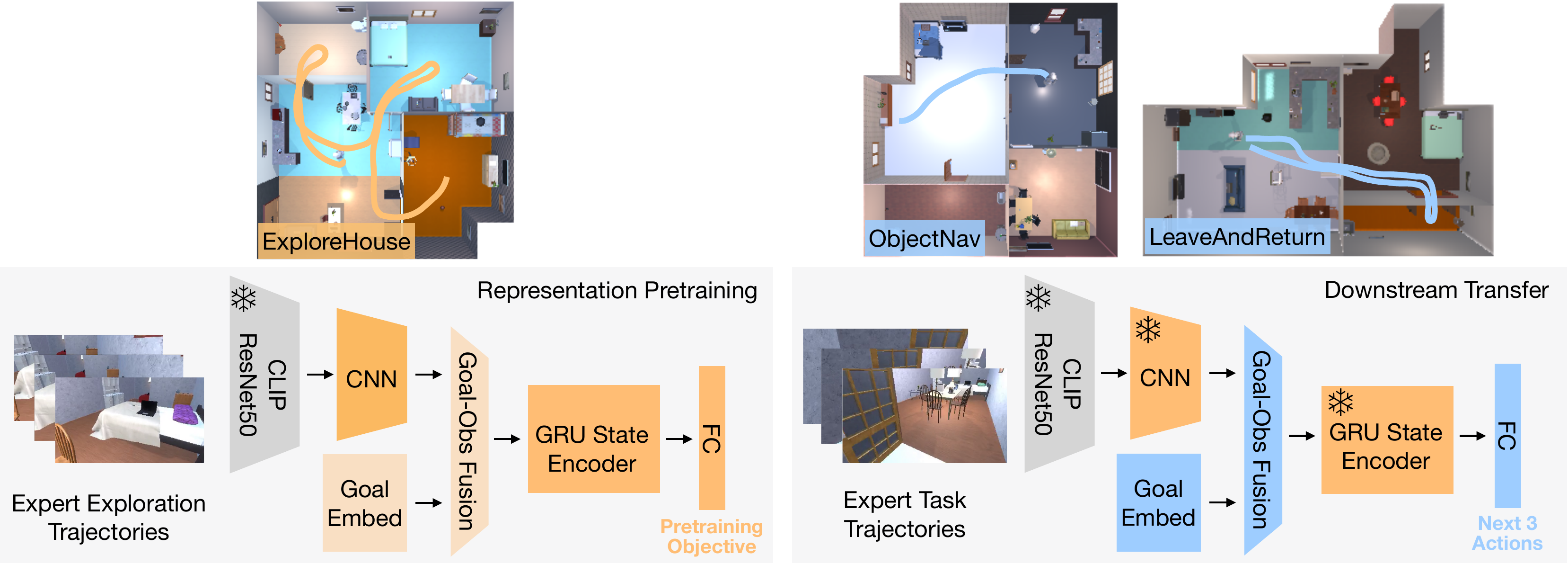}
    \caption{
    Upper, schematic pretraining and evaluation trajectories. Lower, experiment pipeline.  Models receive embeddings of first-person-view RGB images (224$\times$224$\times$3).
    }
    \label{fig:pipeline}
\end{figure}

\subsection{Pretraining}

\textbf{Dataset}. We use a set of high-quality exploration trajectories (\textsc{ExploreHouse}) to pretrain both visual and belief representations, to ensure that a range of interesting behaviors is available during pretraining.  In each episode, an expert agent explores every room in a given house by visiting the center of each room (Figure \ref{fig:pipeline}, upper left). The dataset comprises a total of 9.5K houses generated using ProcTHOR~\citep{deitke2022procthor, kolve2017ai2thor}, with approximately 10 episodes in each house.  The action space includes \textit{move ahead}, \textit{rotate right}, \textit{small rotate right}, \textit{rotate left}, \textit{small rotate left}, \textit{move back}, \textit{subdone}, \textit{done}, where \textit{subdone} is used at the end of visiting each room.

\textbf{Agent}.  We adopt the architecture used in \citet{khandelwal2022simple} (Figure \ref{fig:pipeline}, lower left).  All frames are first embedded through a frozen CLIP ResNet-50 model.  The visual compression module takes as input the current frame's CLIP representation
$V\in\mathbb{R}^{2048\times 7 \times7}$ and processes it using a two-layer CNN to form an observation embedding $O\in\mathbb{R}^{32\times7\times7}$. $O$ is then flattened into a 1568-dim vector and passed into a 1-layer GRU with 512 hidden units.  If a goal embedding exists for the pretraining objective (\eg hindsight-based objectives, see below), the goal embedding $G\in\mathbb{R}^{32}$ is tiled to be of shape 32$\times$7$\times$7 and concatenated with $O$, then passed through another two-layer CNN to update $O$ to be goal-conditioned, before temporal integration in the state encoder.  

\textbf{Pretraining objectives}.  Our preliminary experiments focused on the following pretraining objectives: 1) four auxiliary tasks commonly used in embodied AI; 2) imitation learning, specifically behavior cloning; 3) two hindsight-based objectives inspired by hindsight experience replay \citep{andrychowicz2017hindsight}.  All pertaining runs were trained with 100M frames of experience, and the checkpoint with the highest performance on 1K validation houses was used for downstream evaluation.

The auxiliary losses we test include temporal distance (TempDist), inverse dynamics (InvDyn), forward dynamics (FwdDyn), and action-conditional contrastive predictive coding (CPC|A-4 and CPC|A-8), based on the formulations in \citet{ye2021auxiliarypoint} and \citet{guo2018neural}. We also examine variants of these standard formulations.  For TempDist, we consider a variant similar to \citet{aytar2018playing}, denoted by SimpleTempDist.  At step $t$, the observation embedding $O_{t-d}$ of a random past frame of distance $d$ is projected to the size of the belief state, then combined with the current belief state $B_t$ by element-wise multiplication. The resulting vector is used as input to a 2-layer MLP with size (128, 1) to predict scalar temporal distance.  For the InvDyn variant (SimpleInvDyn), we use the belief at the current frame $B_t$ and the observation embedding at the next frame $O_{t+1}$ to predict the action $a_t$.  For the FwdDyn variant (SimpleFwdDyn), we use the belief at the current frame $B_t$ and the action $a_t$ as input.  We predict the CLIP features of the next frame $V_{t+1}$ (averaged over the spatial dimensions) for both FwdDyn and SimpleFwdDyn, to prevent collapsing to zero when forward dynamics is the only training objective. In CPC|A, the current belief $B_t$ and future actions $a_t, ..., a_{t+n}$ are used in an auxiliary GRU to roll out a set of auxiliary beliefs, which are then paired with ground-truth observation embeddings to contrast whether each pair corresponded to the same step using binary cross-entropy loss \citep{ye2021auxiliarypoint}.  We include a multi-class variant (CPC|A-softmax) where the beliefs from the auxiliary RNN are dot-producted with the observation embeddings in an all-to-all fashion, before using a softmax multi-class cross-entropy to predict the correct belief-observation pair.

The hindsight-based objectives were designed to encourage self-supervised goal-directed pretraining.  We sample either a future frame (HindsightFutureObs) or a partial future trajectory (HindsightFutureTraj) to serve as the current step's goal, bounded by some distance cap (20 in our experiments).  In HindsightFutureObs, the observation embedding of a future frame $O_{t'}$ is concatenated with the current observation embedding $O_t$ to generate a low-dimensional (8 in our experiments) hindsight goal embedding $G$.  $G$ is then combined with $O_t$ through the 2-layer CNN described above, and the resulting embedding is then given to the state encoder.  The model then uses a linear projection of the belief state at each frame $B_t$ to predict the next action.  In HindsightFutureTraj, the hindsight goal embedding $G$ instead comes from a future trajectory of length $n$.  We use an auxiliary 1-layer GRU of the same size as the agent's state encoder to encode \{$O_t, ..., O_{t+n}$\}, and project the final belief state down to $G$ (also 8-dim).  This goal embedding is used to predict \{$a_t, ..., a_{t+n-1}$\}. 

\subsection{Downstream evaluation}

\textbf{Tasks}.  We evaluate the effectiveness of the pretrained representations using multi-step imitation learning on two tasks, \textsc{ObjectNav} and \textsc{LeaveAndReturn} (Figure \ref{fig:pipeline}, upper right).  For \textsc{ObjectNav}, an expert planner finds one of 95 different target objects in each episode.  For \textsc{LeaveAndReturn}, an expert planner first navigates from location A to B in a house, then returns to within 1 meter of location A before the episode ends.  For each task, we generated approximately 100K episodes for training and 1K episodes for evaluation, with approximately 10 episodes in each unique house.  The action space is the same as in the pretraining experiments except \textit{subdone} is not used in these tasks.

\textbf{Agent}.  The agent architecture in downstream evaluation is very similar to that in the pretraining experiments.  Specificaly, we freeze the 2-layer CNN and the GRU state encoder with pretrained weights, and train the goal representation, goal-observation fusion, as well as the final linear readout anew (Figure \ref{fig:pipeline}, lower right).  The goal embedding module for \textsc{ObjectNav} uses 32-dim learnable embeddings of each object type as the goal $G$.  For \textsc{LeaveAndReturn}, we use an auxiliary GRU of the same shape as the agent's state encoder to encode the first half of the episode (where the agent navigates from location A to B).  This auxiliary GRU takes trajectory observations \{$O_0, ..., O_{t}$\} as input, and its final belief is linearly mapped to a 32-dim goal vector $G$.  In both tasks, $G$ is then fused with the visual observations through a 2-layer CNN similar to the pretraining experiments.  The agent's belief state from the main GRU state encoder is used in a final linear readout of the next three actions starting at the current frame.  Each output unit represents a possible action triplet.  To encourage end-of-episode learning, the output units also include combinations of each move action and \textit{done}, as well as a unit exclusively for the \textit{done} action.  The agent is trained using softmax cross-entropy loss for approximately 50M frames of experience.  We report the maximum evaluation performance from three seeds for each model.

\textbf{Baselines}.  We compare representation evaluation results from pretrained weights to four baselines. Three statistical baselines set our lower bounds, including a predict-move-forward-only baseline (move-ahead) and two conditional, count-based baselines that predict the action triplet that occurred the most frequently following each action (modal | A) or following each unique action triplet (modal | T), where the conditional frequencies are computed on the pretraining exploration trajectories.  Our upper bound is an end-to-end baseline of the same architecture as the other experiments except all modules are trained from random weights.

\section{Results}

\begin{figure}[h]
    \vspace{-5mm}
    \centering
    \includegraphics[width=14cm]{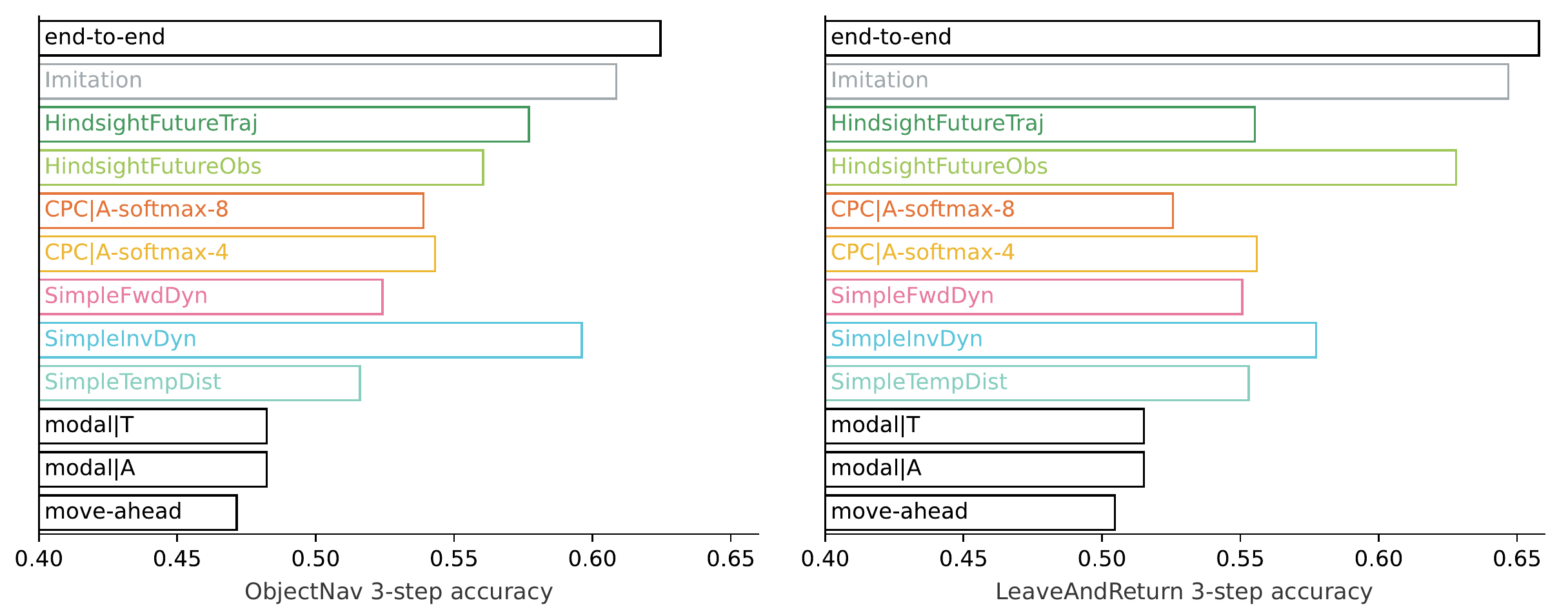}
    \caption{
    Evaluation of the pretrained representations on \textsc{ObjectNav} and \textsc{LeaveAndReturn}.  Each bar shows the best performance among three seeds.
    }
    \label{fig:performance}
\end{figure}

Figure~\ref{fig:performance} shows our preliminary set of multi-step evaluations on different pretrained representations.  Although all pretrained representations were more effective than simple statistical baselines, the ones that were pretrained on single-step action objectives were consistently better than other common auxiliary losses.  To our surprise, even though inverse dynamics and sampling random future frames as goals were sometimes useful for one of the tasks, the pretrained representations from simple imitation learning resulted in the most effective representation transfer to both evaluation tasks.  We note that this resulted from imitating pretraining trajectories that are quite different from downstream tasks (exploration \emph{v.s.} goal-directed shortest path navigation).  These results suggest that imitation learning paired with high-quality pretraining data is an extremely strong baseline, outperforming other, more complex, pretraining objectives with the same data -- importantly, if we do not have high-quality exploration trajectories, imitation would necessarily become much less effective. In the extreme case, we would not expect IL to perform any meaningful representation learning if used with random trajectories whereas other objectives may still result in some learning. A comparative study with random trajectories is left as future work.

A natural question is whether the other more complex action-based pretraining objectives (namely SimpleInvDyn, HindsightFutureObs, and HindsightFutureTraj) were at all effective. We show that hindsight-based approaches and inverse dynamics indeed led to much higher accuracy rates than imitation on pretraining trajectories (Table~\ref{table:pretraining}).  This could imply that, when used as auxiliary objectives in online learning, hindsight-based approaches can potentially still serve as effective learning signals.

\begin{table}[!htb]
  \caption{Single-step action prediction accuracy \emph{on heldout pretraining trajectories} for imitation, inverse dynamics, and hindsight-based objectives.}
  \label{pretrain-action-acc}
  \centering
  \small{
  \begin{tabular}{l|llll}
    \toprule
    Loss & Imitation & SimpleInvDyn & HindsightFutureObs & HindsightFutureTraj \\
    \midrule
    Acc & \hspace{2mm}0.756 & \hspace{5mm}0.867 & \hspace{8mm}0.824 & \hspace{8mm}0.894  \\
    \bottomrule
  \end{tabular}
  }
  \label{table:pretraining}
\end{table}

In addition, we found that auxiliary loss variants generate better representations compared to their original formulations across the board (Table \ref{table:variant}).  In all paired experiments, representations pretrained with the loss variants led to equal or better downstream 3-step accuracy.  For both inverse dynamics and forward dynamics, we found that directly training with the current belief led to better generalization.  The softmax multi-class variant of CPC|A also consistently generalizes better than CPC|A with binary classification.  For temporal distance estimation, we observed that the element-wise multiplication and an additional non-linear layer before scalar distance prediction (as in the SimpleTempDist variant) were key to allowing overfitting on a small training dataset.  Although the resulting representations did not generalize better than TempDist on \textsc{ObjectNav}, its generalization on \textsc{LeaveAndReturn} was better than TempDist by a large margin.

\begin{table}[!htb]
  \caption{Comparing representation transfer of auxiliary loss variants on 3-step imitation.}
  \label{auxloss-variants}
  \centering
  \small{
  \begin{tabular}{llllll}
    \toprule
    Pretrain Trajectories & Downstream Task & Loss & Acc & Loss Variant & Acc  \\
    \toprule
    ExporeHouse & LeaveAndReturn & TempDist & 0.458 & SimpleTempDist & 0.553    \\
    ExporeHouse & ObjectNav & TempDist & 0.518 & SimpleTempDist & 0.516    \\
    ExporeHouse & ObjectNav & InvDyn & 0.558 & SimpleInvDyn & 0.601    \\
    ExporeHouse & ObjectNav & FwdDyn & 0.488 & SimpleFwdDyn & 0.532    \\
    ObjectNav & ObjectNav & CPC|A-4 & 0.515 & CPC|A-softmax-4 & 0.551 \\
    ObjectNav & ObjectNav & CPC|A-8 & 0.516 & CPC|A-softmax-8 & 0.538 \\
    \bottomrule
  \end{tabular}
  }
  \label{table:variant}
\end{table}

\FloatBarrier


\section{Conclusions and Future Work}

Our initial results indicate that imitation learning can serve as a surprisingly strong baseline for this experimental pipeline and outperform the set of common auxiliary and hindsight-based losses we examined.  While the search for useful pretraining objectives remains a crucial direction, these results suggest that the curation of high-quality embodied trajectories that contain interesting behavior may be equally important to help shape useful representations that can generalize to downstream tasks. However, it is possible that imitation performance is not entirely aligned with embodied performance.  As we have only experimented with imitation learning as the evaluation setting so far, we suspect that these preliminary conclusions could change if these pretrained representations are evaluated in online policy learning settings.  This is an important next step for future work.

Although we only experimented with CLIP as the visual backbone, our two-stage evaluation pipeline provides a controlled setting for comparing the pretraining objectives, and it can easily integrate with other visual backbones.  We hypothesize that the relative advantages of these pretraining objectives would hold across different general-purpose visual representations as the observation input, for example, using R3M instead of CLIP as the visual backbone.  Relatedly, as both of our tasks are navigation-based, we are looking to expand the tasks we consider in evaluation, for example, evaluating manipulation-based tasks or established benchmarks (e.g. \citet{majumdar2023we}).  To this end, it is also crucial to consider what metrics would best capture how these representations can support embodied planning.  Our work makes a small step in this direction, focusing on three-step evaluation.  Other metrics, such as a low-data regime, may be needed to gain a holistic view of what representations are considered more effective.

\bibliographystyle{plainnat}

\begin{thebibliography}{16}
\providecommand{\natexlab}[1]{#1}
\providecommand{\url}[1]{\texttt{#1}}
\expandafter\ifx\csname urlstyle\endcsname\relax
  \providecommand{\doi}[1]{doi: #1}\else
  \providecommand{\doi}{doi: \begingroup \urlstyle{rm}\Url}\fi

\bibitem[Andrychowicz et~al.(2017)Andrychowicz, Wolski, Ray, Schneider, Fong, Welinder, McGrew, Tobin, Pieter~Abbeel, and Zaremba]{andrychowicz2017hindsight}
Marcin Andrychowicz, Filip Wolski, Alex Ray, Jonas Schneider, Rachel Fong, Peter Welinder, Bob McGrew, Josh Tobin, OpenAI Pieter~Abbeel, and Wojciech Zaremba.
\newblock Hindsight experience replay.
\newblock \emph{Advances in neural information processing systems}, 30, 2017.

\bibitem[Aytar et~al.(2018)Aytar, Pfaff, Budden, Paine, Wang, and De~Freitas]{aytar2018playing}
Yusuf Aytar, Tobias Pfaff, David Budden, Thomas Paine, Ziyu Wang, and Nando De~Freitas.
\newblock Playing hard exploration games by watching youtube.
\newblock \emph{Advances in neural information processing systems}, 31, 2018.

\bibitem[Batra et~al.(2020)Batra, Gokaslan, Kembhavi, Maksymets, Mottaghi, Savva, Toshev, and Wijmans]{batra2020objectnav}
Dhruv Batra, Aaron Gokaslan, Aniruddha Kembhavi, Oleksandr Maksymets, Roozbeh Mottaghi, Manolis Savva, Alexander Toshev, and Erik Wijmans.
\newblock Objectnav revisited: On evaluation of embodied agents navigating to objects.
\newblock \emph{arXiv preprint arXiv:2006.13171}, 2020.

\bibitem[Deitke et~al.(2022)Deitke, VanderBilt, Herrasti, Weihs, Ehsani, Salvador, Han, Kolve, Kembhavi, and Mottaghi]{deitke2022procthor}
Matt Deitke, Eli VanderBilt, Alvaro Herrasti, Luca Weihs, Kiana Ehsani, Jordi Salvador, Winson Han, Eric Kolve, Aniruddha Kembhavi, and Roozbeh Mottaghi.
\newblock Procthor: Large-scale embodied {AI} using procedural generation.
\newblock \emph{Advances in Neural Information Processing Systems}, 35, 2022.

\bibitem[Guo et~al.(2018)Guo, Azar, Piot, Pires, and Munos]{guo2018neural}
Zhaohan~Daniel Guo, Mohammad~Gheshlaghi Azar, Bilal Piot, Bernardo~A Pires, and R{\'e}mi Munos.
\newblock Neural predictive belief representations.
\newblock \emph{arXiv preprint arXiv:1811.06407}, 2018.

\bibitem[Khandelwal et~al.(2022)Khandelwal, Weihs, Mottaghi, and Kembhavi]{khandelwal2022simple}
Apoorv Khandelwal, Luca Weihs, Roozbeh Mottaghi, and Aniruddha Kembhavi.
\newblock Simple but effective: Clip embeddings for embodied ai.
\newblock In \emph{Proceedings of the IEEE/CVF Conference on Computer Vision and Pattern Recognition}, pages 14829--14838, 2022.

\bibitem[Kolve et~al.(2017)Kolve, Mottaghi, Han, VanderBilt, Weihs, Herrasti, Gordon, Zhu, Gupta, and Farhadi]{kolve2017ai2thor}
Eric Kolve, Roozbeh Mottaghi, Winson Han, Eli VanderBilt, Luca Weihs, Alvaro Herrasti, Daniel Gordon, Yuke Zhu, Abhinav Gupta, and Ali Farhadi.
\newblock {AI2-THOR: An Interactive 3D Environment for Visual AI}.
\newblock \emph{arXiv}, 2017.

\bibitem[Majumdar et~al.(2023)Majumdar, Yadav, Arnaud, Ma, Chen, Silwal, Jain, Berges, Abbeel, Malik, et~al.]{majumdar2023we}
Arjun Majumdar, Karmesh Yadav, Sergio Arnaud, Yecheng~Jason Ma, Claire Chen, Sneha Silwal, Aryan Jain, Vincent-Pierre Berges, Pieter Abbeel, Jitendra Malik, et~al.
\newblock Where are we in the search for an artificial visual cortex for embodied intelligence?
\newblock \emph{arXiv preprint arXiv:2303.18240}, 2023.

\bibitem[Nair et~al.(2022)Nair, Rajeswaran, Kumar, Finn, and Gupta]{nair2022r3m}
Suraj Nair, Aravind Rajeswaran, Vikash Kumar, Chelsea Finn, and Abhinav Gupta.
\newblock R3m: A universal visual representation for robot manipulation.
\newblock \emph{arXiv preprint arXiv:2203.12601}, 2022.

\bibitem[Oquab et~al.(2023)Oquab, Darcet, Moutakanni, Vo, Szafraniec, Khalidov, Fernandez, Haziza, Massa, El-Nouby, et~al.]{oquab2023dinov2}
Maxime Oquab, Timoth{\'e}e Darcet, Th{\'e}o Moutakanni, Huy Vo, Marc Szafraniec, Vasil Khalidov, Pierre Fernandez, Daniel Haziza, Francisco Massa, Alaaeldin El-Nouby, et~al.
\newblock Dinov2: Learning robust visual features without supervision.
\newblock \emph{arXiv preprint arXiv:2304.07193}, 2023.

\bibitem[Radford et~al.(2021)Radford, Kim, Hallacy, Ramesh, Goh, Agarwal, Sastry, Askell, Mishkin, Clark, et~al.]{radford2021learning}
Alec Radford, Jong~Wook Kim, Chris Hallacy, Aditya Ramesh, Gabriel Goh, Sandhini Agarwal, Girish Sastry, Amanda Askell, Pamela Mishkin, Jack Clark, et~al.
\newblock Learning transferable visual models from natural language supervision.
\newblock In \emph{International conference on machine learning}, pages 8748--8763. PMLR, 2021.

\bibitem[Radosavovic et~al.(2023)Radosavovic, Xiao, James, Abbeel, Malik, and Darrell]{radosavovic2023real}
Ilija Radosavovic, Tete Xiao, Stephen James, Pieter Abbeel, Jitendra Malik, and Trevor Darrell.
\newblock Real-world robot learning with masked visual pre-training.
\newblock In \emph{Conference on Robot Learning}, pages 416--426. PMLR, 2023.

\bibitem[Sharma et~al.(2023)Sharma, Fantacci, Zhou, Koppula, Heess, Scholz, and Aytar]{sharma2023lossless}
Mohit Sharma, Claudio Fantacci, Yuxiang Zhou, Skanda Koppula, Nicolas Heess, Jon Scholz, and Yusuf Aytar.
\newblock Lossless adaptation of pretrained vision models for robotic manipulation.
\newblock \emph{arXiv preprint arXiv:2304.06600}, 2023.

\bibitem[Singh et~al.(2022)Singh, Salvador, Weihs, and Kembhavi]{singh2022general}
Kunal~Pratap Singh, Jordi Salvador, Luca Weihs, and Aniruddha Kembhavi.
\newblock A general purpose supervisory signal for embodied agents.
\newblock \emph{arXiv preprint arXiv:2212.01186}, 2022.

\bibitem[Ye et~al.(2021{\natexlab{a}})Ye, Batra, Das, and Wijmans]{ye2021auxiliaryobject}
Joel Ye, Dhruv Batra, Abhishek Das, and Erik Wijmans.
\newblock Auxiliary tasks and exploration enable objectgoal navigation.
\newblock In \emph{Proceedings of the IEEE/CVF International Conference on Computer Vision}, pages 16117--16126, 2021{\natexlab{a}}.

\bibitem[Ye et~al.(2021{\natexlab{b}})Ye, Batra, Wijmans, and Das]{ye2021auxiliarypoint}
Joel Ye, Dhruv Batra, Erik Wijmans, and Abhishek Das.
\newblock Auxiliary tasks speed up learning point goal navigation.
\newblock In \emph{Conference on Robot Learning}, pages 498--516. PMLR, 2021{\natexlab{b}}.

\end{thebibliography}


\end{document}